\documentclass{article}
\usepackage{ijcai24}

\usepackage{times}
\usepackage{soul}
\usepackage{url}
\usepackage[hidelinks]{hyperref}
\usepackage[utf8]{inputenc}
\usepackage[small]{caption}
\usepackage{graphicx}
\usepackage{amsmath}
\usepackage{amsthm}
\usepackage{booktabs}
\usepackage{algorithm}
\usepackage{algorithmic}
\usepackage[switch]{lineno}
\usepackage{amssymb}
\usepackage{xspace}
\usepackage[capitalize]{cleveref}

\newcommand{\etal}{\textit{et al.}\xspace}
\title{Towards Highly Realistic Artistic Style Transfer via Stable Diffusion with Step-aware and Layer-aware Prompt}

\author{
Zhanjie Zhang$^{1}$
\and
Quanwei Zhang$^{1}$\and
Huaizhong Lin$^{1\dagger}$\and
Wei Xing$^{1\dagger}$
Juncheng Mo$^1$
\and
Shuaicheng Huang$^2$\and
Jinheng Xie$^{3}$\and
Guangyuan Li$^1$\and
Junsheng Luan$^1$
\and
 Lei Zhao$^{1\dagger}$\and
Dalong Zhang$^{1}$\and
Lixia Chen$^4$
\\
\affiliations
$^1$Zhejiang University\and
$^2$Anhui University\and
$^3$National Univeristy of Singapore\\
$^4$Zhejiang Gongshang University\\
\emails
\{cszzj, cszqw, linhz, wxing, csmjc, cslgy, l.junsheng121,cszhl\}@zju.edu.cn,\\
Z02114187@stu.ahu.edu.cn,xiejinheng2020@email.szu.edu.cn,
zdlxing@126.com,clixia@163.com
}

\begin{document}

\twocolumn[{%
	\renewcommand\twocolumn[1][]{#1}%
	\maketitle
	\begin{center}
		\centering
		\includegraphics[width=1\linewidth]{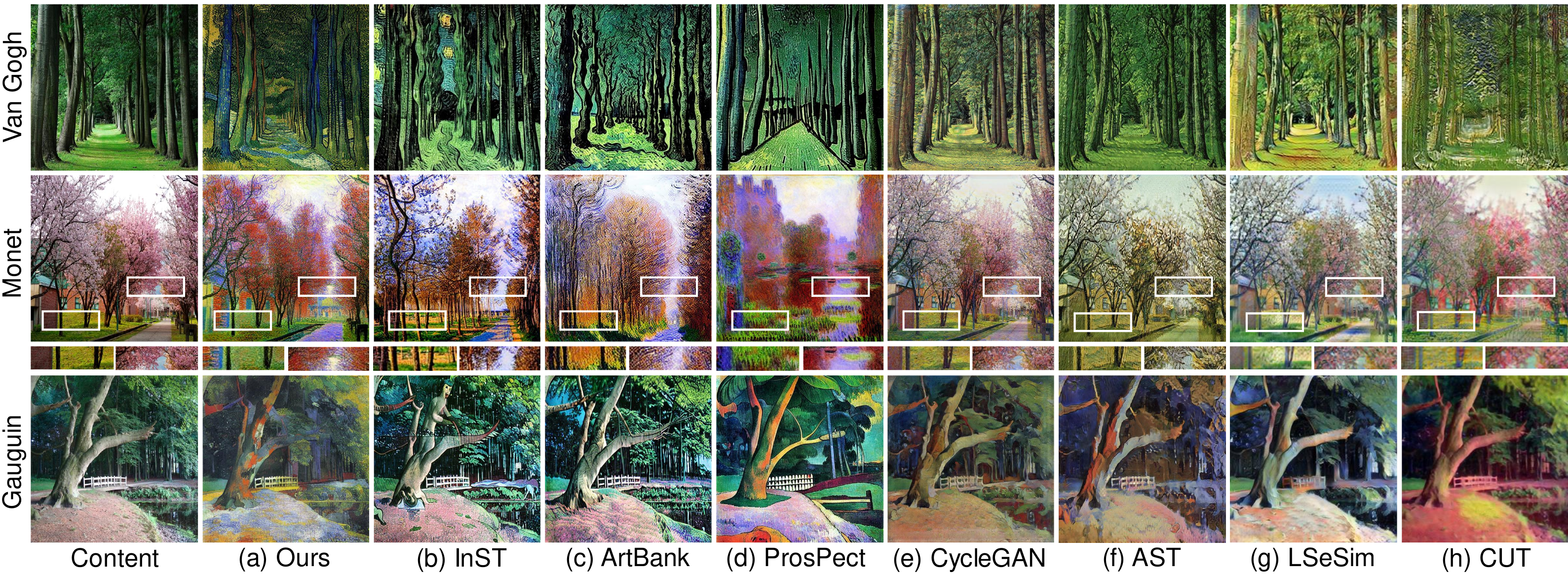}
		\captionof{figure}{Stylization examples with three different styles (i.e., Van Gogh, Monet, Gauguin). Compared to existing state-of-the-art large-scale pre-trained diffusion model-based methods (b-d) and generative adversarial network-based methods (e-h), our proposed method (a) generates highly realistic artistic stylized images and preserves the content structure of input content images well.}
		\label{teaser}
	\end{center} 
}]

\begin{abstract}
	Artistic style transfer aims to transfer the learned artistic style onto an arbitrary content image, generating artistic stylized images. Existing generative adversarial network-based methods fail to generate highly realistic stylized images and always introduce obvious artifacts and disharmonious patterns. Recently, large-scale pre-trained diffusion models opened up a new way for generating highly realistic artistic stylized images. However, diffusion model-based methods generally fail to preserve the content structure of input content images well, introducing some undesired content structure and style patterns. To address the above problems, we propose a novel pre-trained diffusion-based artistic style transfer method, called LSAST, which can generate highly realistic artistic stylized images while preserving the content structure of input content images well, without bringing obvious artifacts and disharmonious style patterns. Specifically, we introduce a Step-aware and Layer-aware Prompt Space, a set of learnable prompts, which can learn the style information from the collection of artworks and dynamically adjusts the input images' content structure and style pattern.
	To train our prompt space, we propose a novel inversion method, called Step-ware and Layer-aware Prompt Inversion, which allows the prompt space to learn the style information of the artworks collection.
	In addition, we inject a pre-trained conditional branch of ControlNet into our LSAST, which further improved our framework's ability to maintain content structure.
	Extensive experiments demonstrate that our proposed method can generate more highly realistic artistic stylized images than the state-of-the-art artistic style transfer methods.
	

\end{abstract}

\section{Introduction}
Artistic style transfer has recently attracted widespread attention in academia and industry since the seminal work of CycleGAN~\cite{zhu2017unpaired}. Existing artistic style transfer methods can be divided into generative adversarial network-based approaches (GAN-based approaches) and large-scale pre-trained diffusion model-based approaches (Diffusion-based approaches). 

More specifically, GAN-based methods~\cite{zhu2017unpaired,kim2017learning,sanakoyeu2018style,kim2019u,park2020contrastive,wang2022aesust} generally utilize generative adversarial network and a training set of aligned/unaligned image pairs to learn the mapping between an input image and an output image. For example, Zhu et al.~\cite{zhu2017unpaired} used two mirror generative adversarial network to learn and improve the mapping between the input image and output image, synthesizing artistic stylized images. Sanakoyeu~\etal~\cite{sanakoyeu2018style} proposed a style-aware content loss, to improve the quality of artistic stylized images by capturing how style patterns affect content structure. However, GAN-based methods are limited by the instability of adversarial training and the scarcity of training data, failing to generate highly realistic artistic stylized images and introducing the obvious artifacts and disharmonious patterns on the stylized images (Please see in Fig.~\ref{teaser} (e-h)).

Large-scale pre-trained diffusion model-based approaches~\cite{dhariwal2021diffusion,huang2022draw,nichol2021glide,wu2022creative,ho2020denoising,hu2023phasic} use massive parameters to learn and store the information from the large-scale training data, possessing the ability to generate highly realistic images. This opens up a new way for generating highly realistic artistic stylized images. For example, Zhang~\etal~\cite{zhang2023artbank} proposed a global prompt space, a learnable parameter matrix, to learn and store the style information from the collection of artworks and condition pre-trained large-scale diffusion model to generate artistic stylized image. Zhang~\etal~\cite{zhang2023prospect} introduced a step-aware prompt space, a set of learnable parameter matrixes, for the whole diffusion process to generate desired stylized image. Although these approaches could generate highly realistic stylizied images, they failed to preserve the content structure of input content image well, bringing some undesired content structure and style patterns (Please see in Fig.~\ref{teaser} (b-d)).

Motivated by the above observation, we propose a novel framework, called LSAST, which can dig out prior knowledge from the large-scale pre-trained diffusion model to generate highly realistic artistic stylized images while preserving the content structure of the input content image well. Specifically, we propose a Step-aware and Layer-aware Prompt Space $P^{*}$ (i.e., a set of learnable prompts) to control the whole diffusion process from both step and layer dimensions, dynamically adjusting the input images' content structure and style pattern. In detail, to make $P^{*}$ layer-aware, we divide Unet into three layers and design different prompts for different layers to control content structure and style information at different scales. To make $P^{*}$ step-aware, we divide the entire de-noising phase into 10 stages, each containing 100 steps, and each stage corresponds to a prompt, which dynamically adjusts the content structure and style information based on the de-noising steps. To train $P^{*}$, we propose a novel inversion method, called Step-ware and Layer-aware Prompt Inversion, which allows the prompt $P^{*}$ to learn the style information of the artworks collection.
Finally, we inject a pre-trained conditional branch of ControlNet~\cite{zhang2023adding} into our LSAST, which further improved our framework's ability to maintain content structure.
 
To summarize, the main contribution of this paper is as follows:
\begin{itemize}
	\item We propose a novel pre-trained diffusion-based artistic style transfer framework, which can generate highly realistic stylized images and preserve the content structure of input content images well without introducing obvious artifacts and disharmonious style patterns.
	\item We design a novel Step-aware and Layer-aware Prompt Space and conduct a Step-aware and Layer-aware Prompt Inversion to train prompt space, which can learn the style information from the collection of artworks and dynamically adjusts the input images' content structure and style pattern.
	\item Extensive quantitative and qualitative experiments demonstrate that our proposed LSAST outperforms the state-of-the-art GAN-based and Diffusion-based methods.
\end{itemize}

\section{Related Work}
\textbf{Generative Adversarial Network-based Methods.} 
Generative adversarial network-based methods refer to the use of discriminator networks~\cite {goodfellow2014generative} to train a well-designed forward network that can bridge a mapping between the input image and output image. As the seminal work of artistic style transfer, Zhu~\etal~\cite{zhu2017unpaired} expanded adversarial loss and proposed Cycle-Consistent loss to improve the quality of artistic stylized images. The cycle-consistent loss inspired a lot of researchers to explore a more effective way to enhance further the quality of artistic stylized images, including~\cite{sanakoyeu2018style,wang2022aesust,jiang2023scenimefy,park2020contrastive,zheng2021spatially,chen2021artistic,fu2019geometry,zuo2023generative,zhang2023caster,zhang2021generating}. For example, Park~\etal~\cite{park2020contrastive} proposed a method to preserve the content structure of input content image by maximizing the mutual information between input image and output image via contrastive learning. Chen~\etal~\cite{chen2021artistic} introduced two contrastive losses to learn internal-external style information, making the color distributions and texture patterns in the stylized image more reasonable and harmonious. Sanakoyeu~\etal~\cite{sanakoyeu2018style} proposed a style-aware content loss to improve the stylization of images by capturing how style affects content. Zheng~\etal~\cite{zheng2021spatially} exploited the spatial patterns of self-similarity to capture spatial relationships within an image rather than domain appearance to preserve the content structure. While generative adversarial network-based methods effectively transfer the learned style onto an arbitrary content image, they fail to generate highly realistic artistic stylized images, introducing obvious artifacts and disharmonious patterns. 

\begin{figure*}[htb]	
	\centering
	\includegraphics[width=1\textwidth,height=0.76\textwidth]{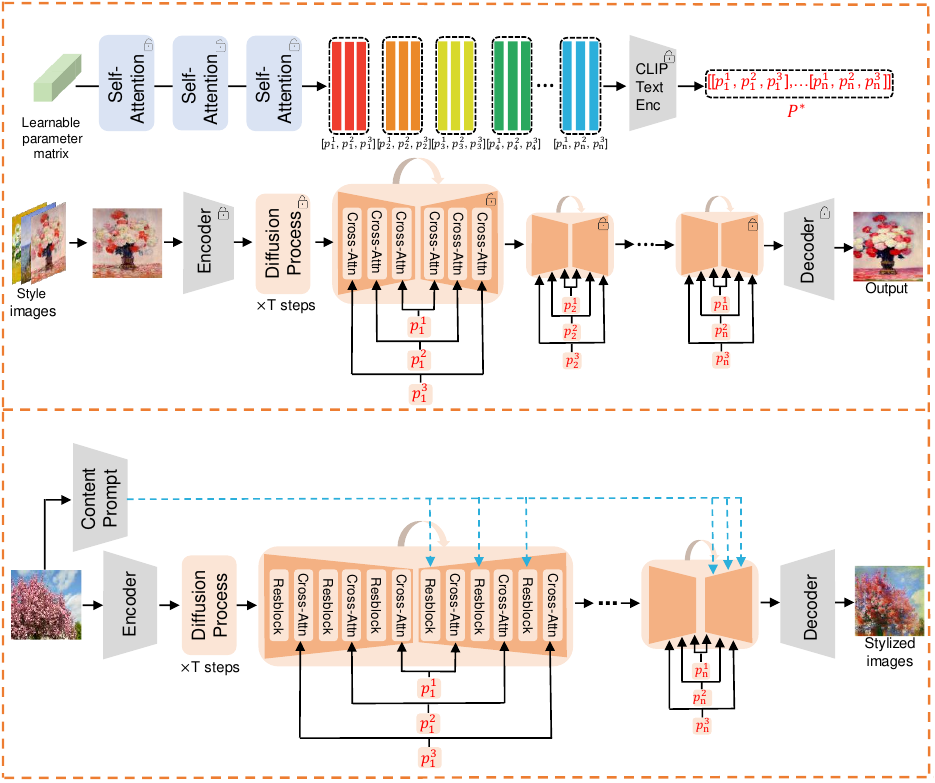} 
	\caption{The overview of our proposed framwork which consists of a training stage and an inference stage. In the training stage, the Step-aware and Layer-aware Prompt Spaces will learn and store the style information from the collection of artworks. In the inference stage, we utilize a pre-trained conditional branch of ControlNet, as content prompt, into our LSAST, which further improved our framework's ability to maintain content structure.
	}
	\setlength{\belowcaptionskip}{-200cm}  
	\label{image4}
\end{figure*}

\textbf{Large-sacle Pre-trained Diffusion Model-based Approaches.} Large-scale pre-trained diffusion model-based approaches use massive parameters to learn information from large-scale data. To dig out the prior knowledge from the pre-trained diffusion model,
Zhang~\etal~\cite{zhang2022inversion} proposed to learn a word embedding from a single style image and then guide the whole diffusion process to transfer learned style information onto a content image. Kim~\etal~\cite{kim2022diffusionclip} performed text-driven image manipulation to preserve the content structure of input content image and learn style information from style image.   Zhang~\etal~\cite{zhang2023artbank} proposed a global prompt condition (i.e., a learnable parameter matrix) to learn and store the style information and condition pre-trained large-scale diffusion model to generate artistic stylized image. Zhang~\etal~\cite{zhang2023prospect} introduced a step-aware prompt condition for the whole diffusion process to generate the stylized image. Xie~\etal~\cite{xie2023boxdiff} proposed to control object synthesis in the given spatial conditions. 
While the large-scale pre-trained model-based approaches can generate highly realistic stylized images, they always fail to preserve the content structure of the input content image well, bringing some undesired content structure and style patterns.


\section{Proposed Method}
Let $I_{s}$ and $I_{c}$ be the style image and content image respectively; our goal is to train a Step-aware and Layer-aware Prompt Space, a set of learnable prompts, to learn the style information from the style images and dig out the abundant prior knowledge from large-scale pre-trained diffusion model to transfer the learned style onto the content image $I_{c}$, synthesizing highly realistic stylized images $I_{cs}$. The pipeline of our proposed LSAST is shown in Fig.~\ref{image4}, which consists of two stages: Learning step-aware and layer-aware prompt space from the collection of artworks (training) and generating highly realistic artistic stylized images (inference).

In the training stage, we utilize a learnable parameter matrix to learn and store the style information from the collection of artworks. Then, we expand it into a set of learnable parameter matrixes (Step-aware and Layer-aware Prompts Space). These learnable parameter matrixes dig out the prior knowledge of the pre-trained diffusion model from step and layer dimensions.

In the inference stage, the Step-aware and Layer-aware Prompts Space (i.e., a set of parameters trained in the training phase) are used to guide the pre-trained diffusion model to transfer the learned style onto the content images, generating highly realistic artistic stylized images. Besides, we use the conditional branch of ControlNet~\cite{zhang2023adding} to extract the content structure of input content image as content prompt and guide pre-trained stable diffusion to preserve the content structure.
\begin{figure}[t]
	\centering
	\includegraphics[width=1\columnwidth]{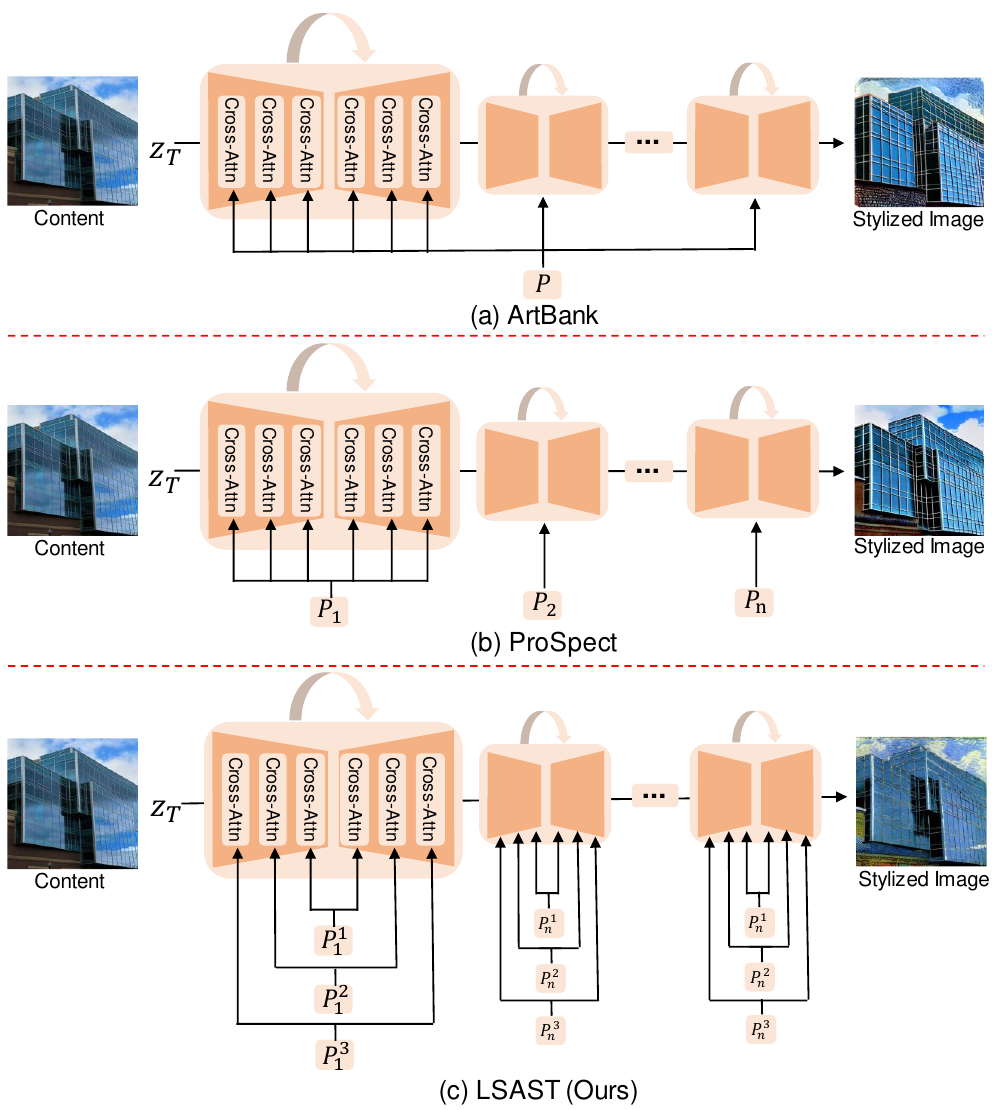} 
	\caption{Differences between (a) ArtBank. (b) ProSpect. (c) LSAST (Ours). Unlike existing methods that use only a global/step-aware prompt to condition the whole diffusion process, LSAST utilizes a step-aware and layer-aware prompt to dynamically adjusts the input images' content structure and style pattern.}
	\label{image2}
\end{figure}

\begin{figure*}[htb]	
	\centering
	\includegraphics[width=1\textwidth,height=0.8\textwidth]{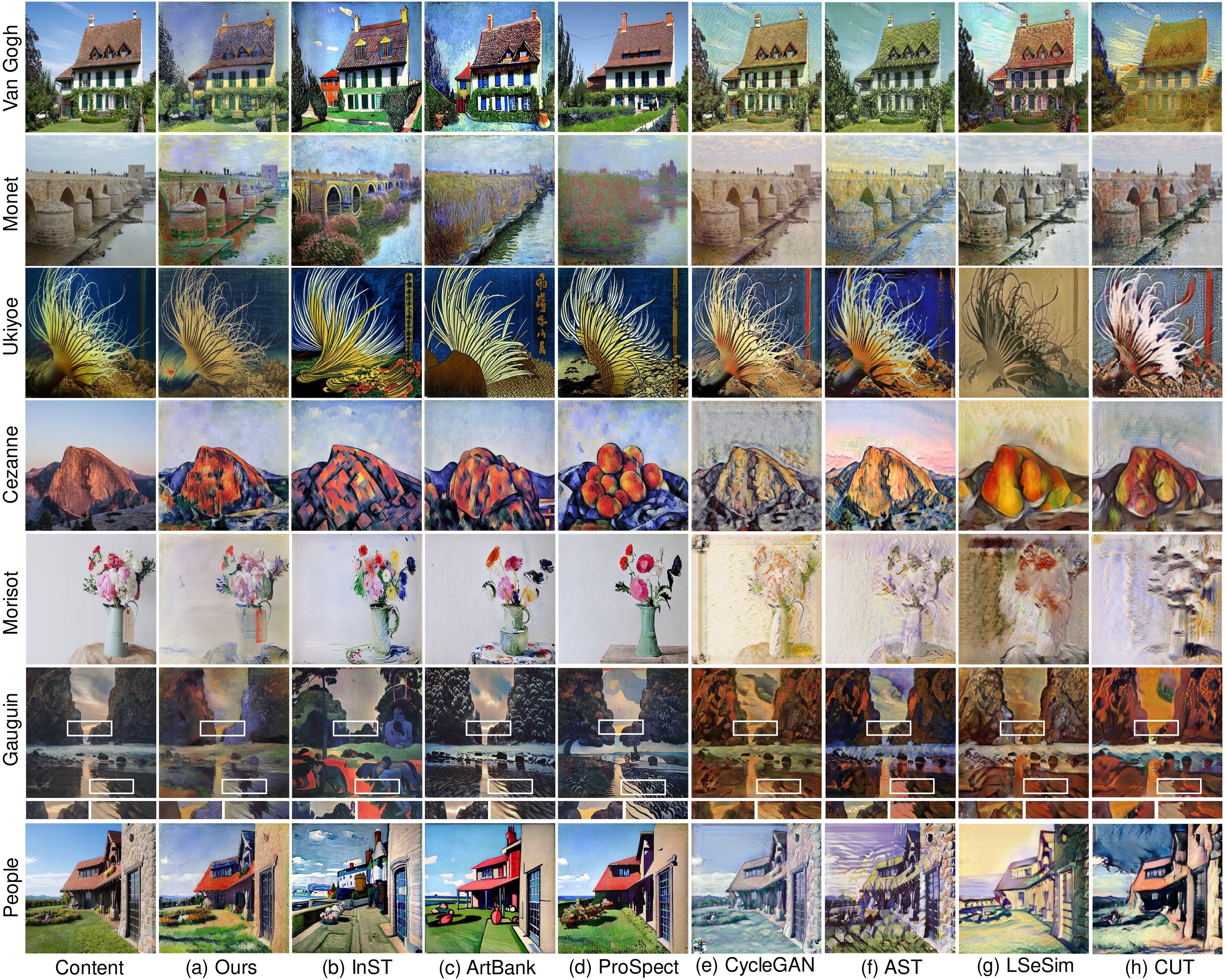} 
	\caption{Qualitative comparisons with other state-of-the-art artistic style methods. The first column shows the input content image. The (b-d) column shows the stylized image from large-scale pre-trained diffusion model-based methods, and the (e-h) column presents the stylized images generated by generative adversarial network-based methods.}
	\setlength{\belowcaptionskip}{-200cm}  
	\label{image3}
\end{figure*}

\subsection{Step-aware and Layer-aware Prompt Space}
Our goal is to dig out the abundant prior knowledge from the large-scale pre-trained Stable Diffusion~\cite{rombach2022high}, which can generate highly realistic images. To dig out the prior knowledge from it, Gal~\etal~\cite{gal2022image} proposed to optimize the token prompt embedding space of CLIP~\cite{radford2021learning}, outputting a suitable global prompt for the diffusion model. However, the denoising process within Stable Diffusion generally takes 1,000 steps.
In the denoising process's early, middle, and late stages~\cite{agarwal2023image}, the text prompt controls the coarse content structure, detail content structure, and detail style patterns, respectively. Only using a global prompt embedding has difficulty in controlling content structure and style patterns simultaneously. Zhang~\etal~\cite{zhang2023prospect} divided the 1,000 denoising steps into ten stages on average, and each stage corresponds to a unique prompt. However, the typical U-Net model, within stable diffusion, can be divided into three layers~\cite{agarwal2023image}: coarse, moderate, and fine, corresponding to coarse, detailed, and style patterns. Thus, only considering dividing the steps into ten stages is still not enough. To this end, we introduce a Step-aware and Layer-aware Prompt Space, which divides the 1,000 denoising steps into ten stages and the U-Net into three layers. Thus, to condition the pre-trained diffusion model from step and layer dimensions, the number of the learnable parameter matrix should be $30$. To this end, we expand a learnable parameter matrix $P$ (i.e., $P$$\in$$R^{1\times 1\times 768}$) into Step-aware and Layer-aware Prompt Space ${P}^*$ (i.e., $P^*$$\in$$R^{30\times 1\times 768}$).
Specifically, we formulate $Q$~(query), $K$~(key) and  $V$~(value) as:
\begin{equation}
\begin{aligned}
Q & =f\left({Norm}\left(P\right)\right), \\
K & =g\left({Norm}\left(P\right)\right), \\
V & =h\left(P\right),\\
\end{aligned}
\end{equation}
where $f$, $g$, and $h$ are 1$\times$1 learnable convolution layers to 
expand the dimension of $P$ from $1\times 1\times 768$ to $30\times 1\times 768$. $Norm$ here denotes channel-wise mean-variance normalization. Then, we calculate attention map $A$ as:
\begin{equation}
\begin{aligned}
A &=\operatorname{Softmax}\left(Q^{\top} \otimes K\right),
\end{aligned}
\end{equation}
where $\otimes$ denotes the matrix multiplication.
\begin{equation}
\begin{aligned}
\widehat{P} &=\left(V\otimes A^{\top}\right),
\end{aligned}
\end{equation}
where $\widehat{P}=[p_1, p_2, \ldots, p_n]$ (i.e., $n=30$). Then, these textual conditions are grouped into ten groups. We refer to grounped textual conditions as Step-aware and Layer-aware Prompt Space ${P}^*$. 
\begin{equation}
\begin{aligned}
{P}^* &=[[p_{1}^{1},p_{1}^{2},p_{1}^{3}],[p_{2}^{1},p_{2}^{2},p_{2}^{3}],\ldots,[p_{n}^{1},p_{n}^{2},p_{n}^{3}]] (n=10),
\end{aligned}
\end{equation}
where $P_{i}^{j}$ denotes the prompt corresponding to $i$-th stage of denoising process and $j$-th layer of U-Net. We define the above process as ${P}^*$ = Self-atten ($P$).
\subsection{Step-aware and Layer-aware Prompt Inversion}
\label{Step-aware and Layer-aware Prompt Inversion}
To dig out knowledge from a large-scale pre-trained diffusion model, a naive method is to unfrozen the whole parameter to learn and store the style information from the collection of artworks. Specifically, we
first add noise $\epsilon \sim \mathcal{N}(0, I)$ onto style images, obtaining the noisy style image $z_{s\_t}$.
Then, feeding the $z_{s\_t}$ into Stable Diffusion and using the following loss:
\begin{equation}
\mathcal{L}_{\text {diff }}=\mathbb{E}_{z, t}\left[\left\|\epsilon-\epsilon_\theta\left(z_{s\_t}, t\right)\right\|_2^2\right],
\end{equation}
Once the $\mathcal{L}_{\text {diff }}$ converges, the fine-tuned Stable Diffusion will possesses the knowledge to generate highly realistic artistic stylized images. However, such a naive way costs extra computation burden. To this end, we propose Step-aware and Layer-aware Prompt Inversion to learn and store the knowledge from the collection of artworks and dig out the abundant prior knowledge from the large-scale pre-trained diffusion model. In detail, 
we feed the $z_{s\_t}$ into pre-trained Stable Diffusion (version 1.5), and train the Step-aware and Layer-aware Prompt Space $P^*$ using the following loss:
\begin{equation}
\mathcal{L}_{\text {p }}=\mathbb{E}_{z, p, t}\left[\left\|\epsilon-\epsilon_\theta\left(z_{s_{-} t},{P}^*, t\right)\right\|_2^2\right],
\end{equation}
Once the $\mathcal{L}_{\text {p }}$ converges, the $P^*$ will store the style information from the collection of artworks and can can be used to condition pre-trained stable diffusion to generate the highly realistic artistic stylized image.
\subsection{Content Prompt}
ControlNet~\cite{zhang2023adding} is widely used to condition large-sacle pre-trained diffusion model to preserve the content structure of input content image. We utilize a aditional branch (Canny edge) of the pre-trained ControlNet  as an aditional content prompt and inject content prompt into pre-trained stable diffusion to further improve the ability to maintain the content structure of the input image

\begin{table*}[htb]
	\caption{Quantitative comparisons. The lower the FID score, the better the image quality. * denotes the average user preference.}
	\centering
	\setlength{\tabcolsep}{0.1cm}
	\begin{center}		
		\begin{tabular}{c|c|ccccccccc}
			\hline
			\footnotesize Metrics&Datasets &Ours&InST&ArtBank&ProsPect& CycleGAN  & AST &LSeSim &CUT 
			\\
			\hline
			\footnotesize  &Van Gogh&\textbf{94.17}&113.26&121.62&118.22&112.04&96.10&105.21&95.61
			\\
			\footnotesize  &Morisot&\textbf{176.70}&221.37&225.47&188.11&204.74&199.78&230.37&201.41
			\\
			\footnotesize  &Ukiyoe &\textbf{92.03}&104.73&111.50&93.15&140.78&148.17&130.77&115.44
			\\
			\footnotesize \normalsize{FID}$\downarrow$& Monet &\textbf{130.25}&159.39&166.70&133.73&170.50&123.17&162.25&150.08
			\\
			\footnotesize &Cezanne &\textbf{130.28}&150.24&143.90&139.51&151.14&133.47&141.33&141.34
			\\
			\footnotesize &Gauguin &\textbf{143.40}&145.65&155.27&125.03&160.00&164.73&172.18&165.69
			\\
			\footnotesize &Peploe &\textbf{158.44}&188.37&182.51&186.45&167.77&222.30&192.88&164.39
			\\
			
			\hline
			\footnotesize Time/sec $\downarrow$&-&4.1325&4.0485&3.7547&3.925&0.0312&0.0312&0.0365&0.0312
			\\
			\footnotesize Preference$\uparrow$&-& \textbf{0.634}*&\textbf{0.581}/0.419&\textbf{0.532}/0.468&\textbf{0.545}/0.455&\textbf{0.692}/0.308&\textbf{0.674}/0.326&\textbf{0.689}/0.311&\textbf{0.726}/0.274
			\\
			\footnotesize Deception$\uparrow$&-&\textbf{0.752} &{0.716}&{0.614}&{0.605}&{0.582}&{0.579}&{0.547}&{0.508}
			\\
			\hline 
		\end{tabular}
	\end{center}
	\label{FID}
\end{table*}

\section{Experiments}
\subsection{Implementation Details}
We adopt the large-scale pre-trained stable diffusion model (version 1.5) as our backbone due to its strong ability to generate highly realistic images. We implement our LSAST on two NVIDIA RTX 3090 GPUs via Pytorch~\cite{paszke2019pytorch}. The learning rate is initially set to 0.000005 and then decreased by a factor of 10 after 20,000 iterations (100,000 iterations in total). In training, we collect seven collections of artworks (e.g., Van Gogh, Cezanne, Peploe, Ukiyoe, Monet, Morisot, and Gauguin) from AST~\cite{sanakoyeu2018style} as artistic style images to train our LSAST. Before feeding artistic style images into LSAST, all the style images will be resized to 512$\times$512  pixels.
In inference, we randomly select some content images from DIV2K~\cite{agustsson2017ntire} and resize them to 512$\times$512 as the initial input content image.
\subsection{Qualitative Comparison.} 
We compare our method with the SOTA GAN-based and Diffusion-based artistic style transfer methods as shown in Fig.~\ref{image3}, including InST~\cite{zhang2022inversion}, ArtBank~\cite{zhang2023artbank},
ProsPect~\cite{zhang2023prospect}, AST~\cite{sanakoyeu2018style}, CycleGAN~\cite{zhu2017unpaired}, LSeSim~\cite{zheng2021spatially} and CUT~\cite{park2020contrastive}. As the representative of GAN-based approaches, CycleGAN generally introduces some obvious artifacts into the stylized images (e.g., $4^{th}$ and $5^{th}$ rows). AST sometimes shows blurry content structure and abrupt colors (e.g., $5^{th}$ and $7^{th}$ rows). LSeSim introduces some obvious artifacts and disharmonious patterns (e.g., $5^{th}$ and $6^{th}$ rows). CUT always introduces some unwanted content structure and style patterns (e.g., $5^{th}$ and $7^{th}$ rows). As the representative of Diffusion-based model, the stylized image of InST fails to preserve the content structure of input content image, even deviating from the original content image's semantic (e.g., $6^{th}$ rows). ArtBank generally fails to capture some detailed style patterns (e.g., $6^{th}$ and $7^{th}$ rows). Prospect sometimes has limitations in preserving the content structure of input content image (e.g., $4^{th}$ rows).
In comparison, our proposed LSAST not only can generate highly realistic artistic stylized images without introducing obvious artifacts and disharmonious patterns while preserving the content structure of the input content image well.

\subsection{Quantitative Comparisons}
\textbf{Frechet Inception Distance.} Frechet Inception Distance (FID)~\cite{heusel2017gans} utilizes Inception-V3 model~\cite{szegedy2016rethinking} to extract image features and calculate the distance between generated artistic stylized image's distribution and style image's distribution. The lower FID score means that the distribution of the generated artistic stylized images is closer to that of real artistic images. In other words, the generated artistic stylized image is prone to be regarded as a human-created artistic image. Take Van Gogh's style, for example; we collect 10,000 content images and synthesize 10,000 stylized images for our LSAST and other SOTA artistic style transfer methods to calculate the FID score. As shown in Table.~\ref{FID}, we proposed LSAST to achieve a better FID score, which means our proposed method can generate highly realistic artistic stylized images closer to human-created artistic images.

\begin{figure*}[htb]	
	\centering
	\includegraphics[width=1\textwidth,height=0.37\textwidth]{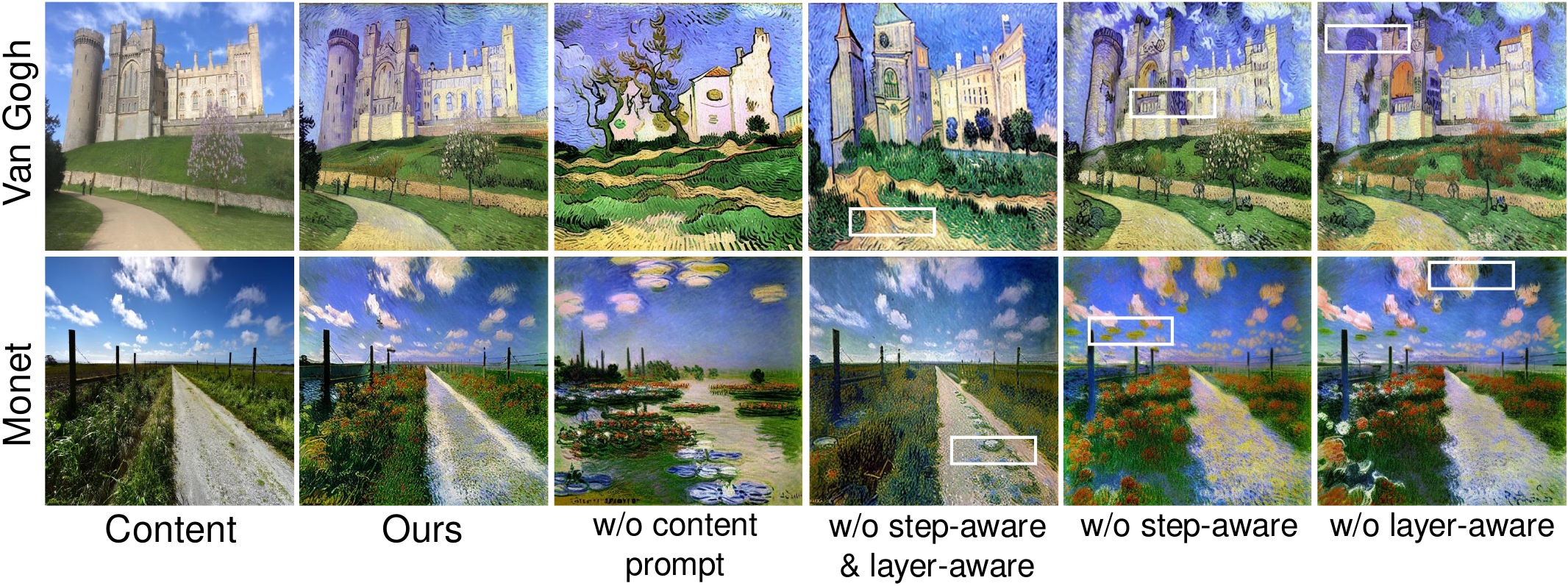} 
	\caption{Ablation studies of our LSAST. Please zoom-in for better comparison.}
	\setlength{\belowcaptionskip}{-200cm}  
	\label{image5}
\end{figure*}
\begin{table*}[htb]
	\caption{The quantitative ablation studies of LSAST. The lower the FID score, the better the image quality.}
	\centering
	\setlength{\tabcolsep}{0.1cm}
	\begin{center}		
		\begin{tabular}{c|c|cccccccccc}
			\hline
			\footnotesize Metric&Datasets&Full Model &w/o content& w/o step\&layer&  w/o step-aware&  w/o layer-aware
			\\
			\hline
			\footnotesize &Van Gogh&\textbf{94.17}&95.05&96.33&94.46&94.58
			\\
			\footnotesize &Morisot&\textbf{176.70}&177.51&178.29&177.54&177.61
			\\
			\footnotesize &Ukiyoe &\textbf{92.03}&93.85&94.97&93.26&93.17
			\\
			\footnotesize \normalsize{FID$\downarrow$}&Monet &\textbf{130.25}&131.78&133.34&130.64&130.72
			\\
			\footnotesize &Cezanne &\textbf{130.28}&131.07&132.84&131.16&131.78
			\\
			\footnotesize &Gauguin &\textbf{143.40}&144.02&145.90&144.27&144.51
			\\
			\footnotesize &Peploe &\textbf{158.44}&159.15&160.32&159.49&159.71
			\\
			\hline 
		\end{tabular}
	\end{center}
	\label{ablation2}
\end{table*}


\textbf{Preference Score.} 
The preference score effectively conducted an A/B Test user study to compare the stylization effects between our method and the other SOTA method, reflecting the popularity of our method compared to other state-of-the-art methods. We randomly choose 100 content images to synthesize 100 stylized images for each method. Then, the content image and corresponding two stylized image pairs generated by our method and a randomly selected SOTA are displayed. The subjects need to choose their preferred stylization effections. We collected 5000 votes from 50 subjects and showed the preference score in Table.~\ref{FID} ("Preference" row). The results indicate that our LSAST achieves the best popularity.

\textbf{Deception Score.} 
The deception score is an effective means for evaluating whether artistic stylized images are prone to be regarded as human-created from people's perspective. A higher deception score means the artistic stylized image is more prone to be considered a human-created artistic image. To compute the deception score, we first randomly choose 20 content images. Then, we feed these content images into our proposed LSAST and other SOTA artistic style transfer methods, generating stylized images. We asked 50 subjects to distinguish whether the generated image was human-created. The result is shown in Tab.~\ref{FID}. A higher deception score means the stylized images are more likely to be considered as human-created artistic images. Besides, we report the deception score of the human-created artistic style image, chosen randomly from WikiArt~\cite{wikiart} is 0.863. From Tab.~\ref{FID}, the deception of our method is closer to 0.863, which indicates that our LSAST achieves a better deception score.

\textbf{Time Information.}
The ``Time/sec" row of Tab.~\ref{FID} shows the inference time comparison on images with a scale of $512\times 512$ pixels. The GAN-based method is faster than diffusion-based methods because diffusion-based methods commonly possess massive parameters. Our approach is also slower than other diffusion-based approaches due to the additional use of pre-trained content prompts. Still, it is undeniable that our approach achieves the best results on all other metrics.

\subsection{Abalation Studies}
To verify the effectiveness of our proposed Step-aware and Layer-aware Prompt Inversion, we split it into Step-aware Prompt Inversion and Layer-aware Prompt Inversion.

\textbf{With and without Step-aware Prompt Inversion.} As introduced in Section.~\ref{Step-aware and Layer-aware Prompt Inversion}, the Step-aware Priompt Inversion divides the 1,000 steps of denoising into three stages from step dimension: coarse content structure stage, detailed content structure stage, and detailed style patterns stage. Here, we train an LSAST model that does not involve step-aware prompt inversion. The experimental results are shown in Fig.~\ref{image5}. Compared with the full LSAST model, we can see that the LSAST model without a step-aware prompt introduces some obvious artifacts and disharmonious style patterns into the stylized image. The step-aware prompt inversion is vital to generating highly realistic stylized images.

\textbf{With and without Layer-aware Prompt Inversion.} The Layer-aware Prompt Inversion divides the architecture of U-Net into three layers from layer dimension: coarse content structure layer, detailed content structure layers, and detail style patterns layer. We show stylized images without layer-aware prompt inversion, as shown in Fig.~\ref{image5}. The stylized image shows some obvious artifacts and disharmonious style patterns. 

\textbf{With and without Step-aware and Layer-aware Prompt Inversion.} We train an LSAST model without step-aware and layer-aware prompt inversion. As shown in Fig.~\ref{image5}, the stylized shown shows some unwanted content structure and style patterns.

\textbf{Without content prompt.} If removing the content prompt in inference, the stylized image fails to preserve the content structure of the input content image. 

\section{Conclusion}
In this paper, we propose a novel framework, which can learn the style information from the collection of artworks and dynamically adjusts the input images' content structure and style pattern, generating highly realistic stylized images (i.e., harmonious patterns and detailed texture) and preserving the content structure of the input content image well. 
Extensive quantitative and qualitative experiments verify that our proposed LSAST outperforms the SOTA GAN-based and Diffusion-based methods.

\bibliographystyle{named}
\bibliography{ijcai24}

\end{document}